# Optimization of Infectious Disease Intervention Measures Based on Reinforcement Learning - Empirical analysis based on UK COVID-19 epidemic data

## Author Information

### Affiliations

**College of Computer Science and Electronic Engineering, Hunan University, Changsha 410082, China.**

Baida Zhang & Yakai Chen

**Academy of Military Medical Sciences, Beijing 100071, China**

Huichun Li & Zhenghu Zu

### Contributions

Baida Zhang contributed to conceptual development, provided resources, and edited the manuscript., Yakai Chen performed experiments and wrote the manuscript draft，Huichun Li analysed the results, Zhenghu Zu provided conceptual advice and analysis. All authors contributed to data interpretation and provided feedback on the manuscript. All authors approved the final version of the manuscript.

### Corresponding author

Correspondence to: Baida Zhang & Zhenghu Zu

## Abstract

Globally, the outbreaks of infectious diseases have exerted an extremely profound and severe influence on health security and the economy. During the critical phases of epidemics, devising effective intervention measures poses a significant challenge to both the academic and practical arenas. There is numerous research based on reinforcement learning to optimize intervention measures of infectious diseases. Nevertheless, most of these efforts have been confined within the differential equation based on infectious disease models. Although a limited number of studies have incorporated reinforcement learning methodologies into individual-based infectious disease models, the models employed therein have entailed simplifications and limitations, rendering it incapable of modeling the complexity and dynamics inherent in infectious disease transmission. We establish a decision-making framework based on an individual agent-based transmission model, utilizing reinforcement learning to continuously explore and develop a strategy function. The framework's validity is verified through both experimental and theoretical approaches. Covasim, a detailed and widely used agent-based disease transmission model, was modified to support reinforcement learning research. We conduct an exhaustive exploration of the application efficacy of multiple algorithms across diverse action spaces. Furthermore, we conduct an innovative preliminary theoretical analysis concerning the issue of "time coverage". The results of the experiment robustly validate the effectiveness and feasibility of the methodological framework of this study. The coping strategies gleaned therefrom prove highly efficacious in suppressing the expansion of the epidemic scale and safeguarding the stability of the economic system, thereby providing crucial reference perspectives for the formulation of global public health security strategies.

## Introduction

In recent years, the outbreak of COVID-19 has demonstrated the colossal destructive potential of infectious diseases[1]. It has inflicted not only grievous harm upon human health but has also flung down a stern gauntlet to the global economic edifice and social stability. As per the statistics provided by the World Health Organization, COVID-19 has precipitated the demise of millions of individuals and engendered losses amounting to trillions of US dollars in the global economy[2]. The rapid transmissibility traits and extensive reach manifested by infectious diseases urgently beckon us to pioneer innovative endeavors in the devising of public health policies and the execution of control modalities. Conventional infectious disease prevention and control strategies encompass isolation, vaccination, social distancing impositions, and travel curbs[3]. Although these measures do possess a certain degree of efficacy in some scenarios, their effectiveness is frequently circumscribed under circumstances characterized by resource paucity or fragmentary information[4].

During the outbreak of infectious diseases, decision-makers are in urgent need of formulating control strategies with promptness and precision so as to mitigate the number of infections and fatalities[5]. Traditional decision-making paradigms predominantly relied on expert insights, yet they were unsuitable for addressing the complexities of infectious disease transmission[6]. In a complex system, a minute alteration might precipitate unpredictable and substantial discrepancies in dynamics. Ergo, it is of utmost criticality to adopt data-driven approaches to support decision-making, especially to real-time recalibration the strategy to meet multiple resource constraints.

In the realm of epidemiology, infectious disease transmission models serve as indispensable tools for unraveling and prognosticating the complex dynamics of epidemics. These models can be broadly categorized into two principal classes: the classic differential equation-based models, exemplified by the renowned SIR and SEIR models[7], and the more intricate individual-based system models, such as ABM, FRED[8], and Covasim[9]. The classic models, prized for their simplicity and operational ease, have been extensively employed. However, they encounter limitations when trying to capture the heterogeneous behaviors shown by diverse groups within complex epidemic scenarios. In contrast, agent-based models have garnered increasing attention in recent years. By carefully simulating specific individual behaviors and enabling the dynamic adjustment of multiple intervention measures, they provide higher accuracy in representing epidemics.

Reinforcement learning, a complex machine learning model focused on dynamically figuring out the best strategies, has shown great potential in unstable and uncertain environments[10,11]. In the field of infectious disease prevention and control, RL has been used to improve intervention strategies. These strategies cover vaccine distribution, isolation procedures, and testing and tracing plans. The main goal is to reduce economic and social costs as much as possible while effectively stopping the spread of the epidemic. A lot of reinforcement learning algorithms have been used in related studies[12,13]. These include Deep Q-Network (DQN)[14], Deep Deterministic Policy Gradient (DDPG)[15], and Proximal Policy Optimization (PPO)[16], along with other algorithms (ICM[17], HRL[18], SAC[19], etc.).

With respect to infectious disease control measures, esearch efforts have mainly focused on non-pharmaceutical intervention (NPI) and pharmaceutical intervention (PI) strategies. In the area of non-pharmaceutical intervention, 20 carefully studied the effects of different lockdown strategies. 21 explored the best time to start the intervention. 22 investigated the importance of testing, contact tracing, and isolation. Meanwhile, 23 examined the impact of isolation strategies on disease spread and economic activities. In the field of pharmaceutical intervention, 24 came up with a new model that combines evolutionary strategies to better control the epidemic. However, most of them employed traditional optimization methods rather than intelligent decision-making approaches.

In recent years, the use of RL in optimizing infectious disease interventions has gradually become more popular. It has spread across different research areas, like dynamic isolation strategies, optimizing vaccine distribution[25], and testing and tracing plans.

In the area of differential equation-based models, 26 investigated how to use RL to search for strategies of lockdowns and travel restrictions that can maximize the control of the spread of the epidemic. 27 created the DURLECA framework to simultaneously suppress the spread of the epidemic and maximize the preservation of urban mobility. 28 used PPO algorithm to learn about school closure policies and explore the advantages of inter-regional cooperation. 29 used RL to construct epidemic intervention plans based on the EpiPolicy model. 30 built the EpidRLearn simulator to analyze the role of RL in epidemic control.

In the area of agent-based models, 31,32 created the Intelligent Disease Response and Lockdown Enforcement Control Agent (IDRLECA) and used HRL to control the epidemic through individual mobility interventions. 33 created virtual epidemic scenarios and use the Double Deep Q-Network (DDQN)[34] to learn lockdown strategies. 35 developed a simulator PANDEMICSIMULATOR and utilizes RL algorithms to optimize government's intervention policies.

The infectious disease model selected for this study is Covasim, which has been carefully applied to policy studies in multiple nations. 36 utilized Covasim to appraise control measures in the Seattle area, 37 analyzed the impact of testing and isolation measures in the UK, and 38 investigated optimization strategies for epidemic response in Vietnam. Additionally, 39 proposed a multi-objective optimization framework combining Covasim and reinforcement learning, but they designed the action space as a discrete one and combined different intervention measures into various actions. So, the design of the action space was rough, and the training curve fluctuated greatly.

In summary, both non-pharmaceutical and pharmaceutical intervention measures are important in epidemic control, and reinforcement learning offers a new perspective for dynamic optimization. Although extant studies have shown the potential of reinforcement learning (RL), their action spaces are relatively simple and still have room for optimization. Grounded on the Covasim model, this study tries to further explore the optimization ways for infectious disease intervention strategies, thereby proffering robust support for epidemic control and policy formulation. The contributions of this study are as follows:

1. Based on the agent model, a new framework has been designed for intelligently generating intervention strategies using reinforcement learning.
2. We established a cross-algorithm validation framework that employs DQN and PPO reinforcement learning paradigms to systematically validate operational feasibility and efficiency improvements. This framework benchmarks these improvements against traditional lockdown policies using multi-dimensional evaluation metrics based on UK COVID-19 epidemic data.
3. We pioneer a temporal optimization analysis of interventions to support the dynamic scheduling of policies.

## Results

### Parameter calibration with real-world data

This research aims to apply reinforcement learning techniques to dynamically optimize the parameters of interventions during epidemic outbreaks. Therefore, a fundamental requirement is that the modified model must accurately reflect real-world epidemic dynamics.

To further validate Covasim, the cumulative number of confirmed cases and deaths in the United Kingdom from January 21, 2020, to May 20, 2020, was used to compare with simulation output generated by Covasim (For specific details, see Table 4). Here, the parameters of the initial number of infected individuals and the initial transmission rate were optimized by using the Optuna[40] library to match the real-world data collected. The final fitting result is presented in Figure 1.

**Fig. 1: UK Data Calibration Chart.**

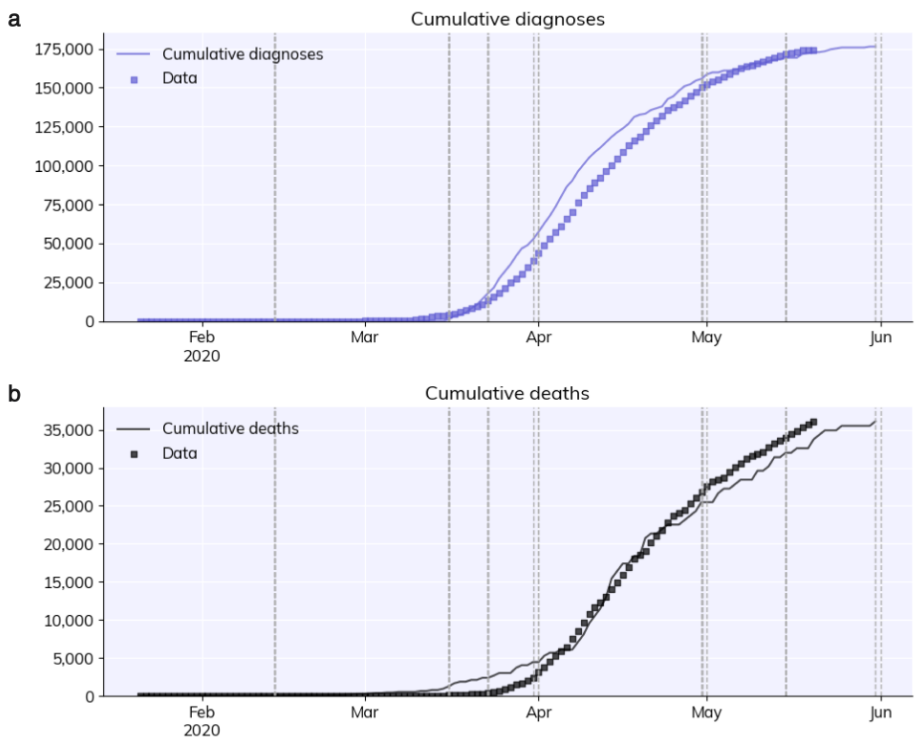

**a** The degree of matching between the cumulative number of diagnosed cases and the real data. **b** The degree of matching between the cumulative number of deaths cases and the real data. The dotted lines in the figure represent the time when various intervention measures were implemented or the intensity of intervention measures were changed, and reflect actual measures implemented by the UK at that time.

As illustrated in Figure 1 that the simulation results after adjusting the parameters show a great match with the real cumulative number of confirmed cases and deaths. This strongly proves the effectiveness of this infectious disease model. Further, Figure 2 depicts the trend chart of epidemic development and provides a clear visualization of how the epidemic would progress.

**Fig. 2: Epidemic Development Trend in the Simulation Environment.**

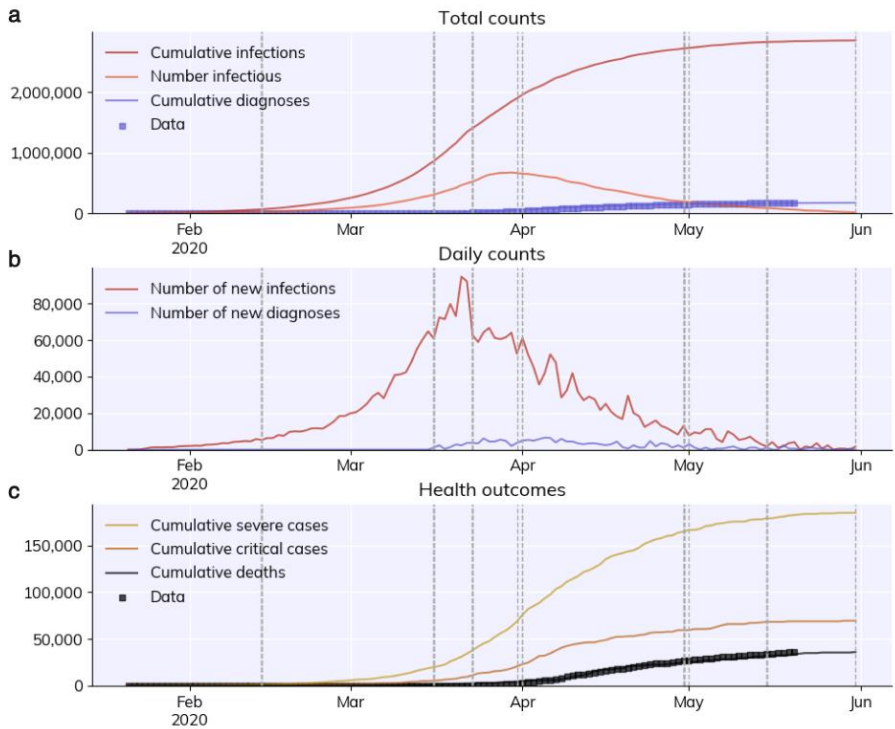

**a** The overall trend. The term "Number Infectious" indicates how many people are in an infectious state per day. **b** The daily numbers of newly infected and diagnosed people. **c** The number of severe cases and deaths.

## Using RL in the Environment Calibrated by Real-world Data

The adjusted parameters along with the basic parameters obtained in the last section can be used as basic data that accurately model the actual epidemic situation in the UK. Then RL was used to optimize the intervention measures.

Firstly, the performance of PPO algorithm and DQN algorithm with prioritized experience replay (PER) in discrete action space were tested. The training results are shown in Figure 3.

**Fig. 3: Comparison of the Training Results of DQN Algorithm with PER and PPO Algorithm in Discrete Action Space.**

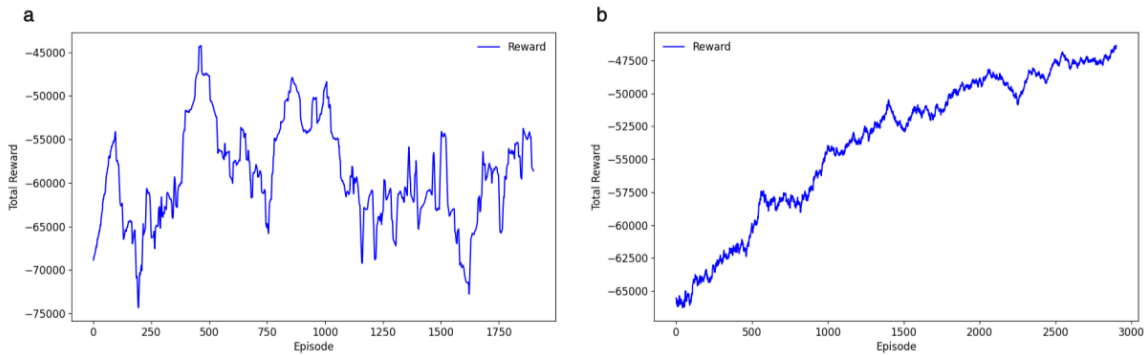

**a** The training result of DQN algorithm. **b** The training result of PPO algorithm. The horizontal axis represents the number of episodes, and the vertical axis represents the total rewards of an episode.

As depicted in Figure 3, when the DQN algorithm with PER is used, it shows significant fluctuations and has a poor convergence effect. In comparison, although the PPO algorithm needs almost 2800 episodes to converge, the convergence return is better than DQN algorithm.

When we look closely at the specific results shown in Figure 4**a** and 4**b**, it is obvious that effective intervention measures can be found with the help of reinforcement learning. Compared with the baseline situation as shown in Figure 2, the strategies learned by both DQN algorithm and PPO algorithm are better at preventing the epidemic than the real strategy. This is specifically shown as a big decrease in the cumulative numbers of infected and dead people. The cumulative number of infected people is limited to about 320,000. However, when looking at the specific action sampling values, we find big differences. Comparing Figure 4**c** and 4**d**, it's clear that the strategy learned by DQN algorithm will carry out an emergency high-lockdown strategy with a high testing and contact tracing measure. But after the emergency strategy, the DQN algorithm strategy only enforces a low intensity lockdown and doesn't carry out testing or tracing measures until the last few weeks. Although it can achieve the goals of controlling the epidemic scale and keeping economic stability, it should be noted that this strategy is limited by the discrete action space settings. If the discrete action space set is changed, the learned strategy will also change. Therefore, it is necessary to set a reasonable discrete action space according to the actual situation. In contrast, the strategy obtained by PPO algorithm shows big swings as shown in Figure 4**d**. Although a high-intensity lockdown will also be implemented in the early stage, along with maintaining a certain level of testing and contact tracing, the strategy has been fluctuating and unstable ever since. The implementation cost of such a policy needs to be carefully considered in the real world. And This indicates potential directions for future improvement.

**Fig. 4: Comparison of Strategies Learned in Discrete Action Space.**

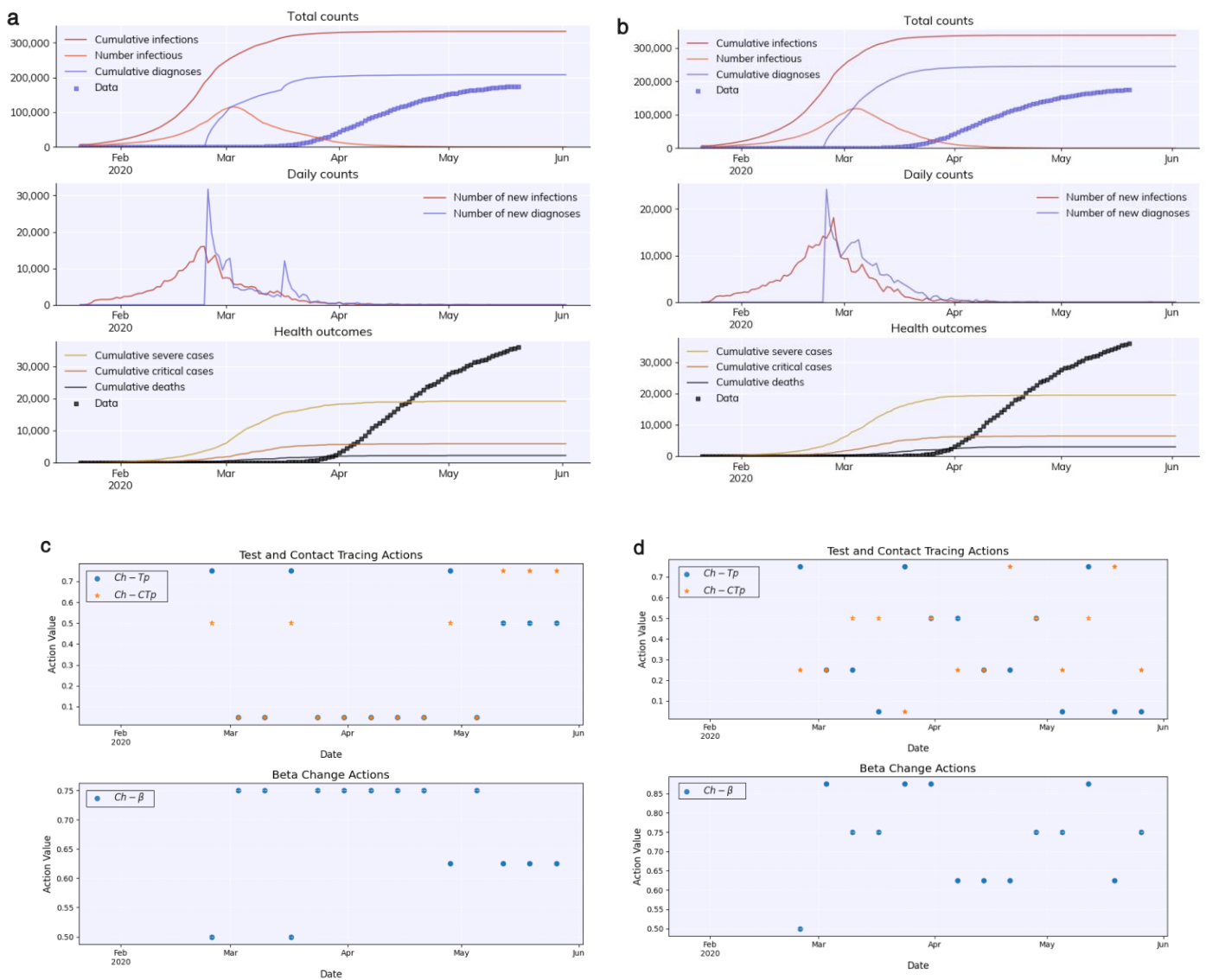

**a** The overall effect diagram of the policy learned by DQN algorithm. **b** The overall effect diagram of the policy learned by PPO algorithm. **c** The sampling diagram of the policy learned by DQN algorithm. **d** The sampling diagram of the policy learned by PPO algorithm

Then, the performance of PPO algorithm in continuous action space were tested. First, the training curve converges stably as shown in Figure 5.

**Fig. 5: Training Results of PPO Algorithm in Continuous Action Space.**

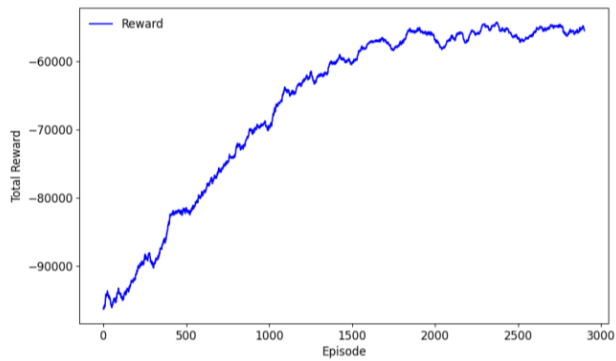

More specifically, the strategy learned by PPO algorithm in continuous action space achieves a better intervention effect, with a cumulative number of infections amounting to only 300,000 depicted in Figure 6**a** and the actions are more reasonable. As shown in Figure 6**b** , as soon as the epidemic is detected, a high intensity lockdown is carried out simultaneously with rigorous testing and contact tracing. Subsequently, the lockdown measure is lifted. Meanwhile, the intensity of testing and contact tracing measures gradually decreases over time until only the contact tracing measure remains in effect. Then to prevent the epidemic from spreading again, a high-intensity lockdown is reintroduced along with a temporary resumption of testing measures. After the epidemic has been brought under better control, only testing and contact tracing maintain alternately. This strategy fully considers the characteristics of epidemic transmission and the impact of intervention measures on the economy.

**Fig. 6: The Strategy Learned by PPO Algorithm in Continuous Action Space.**

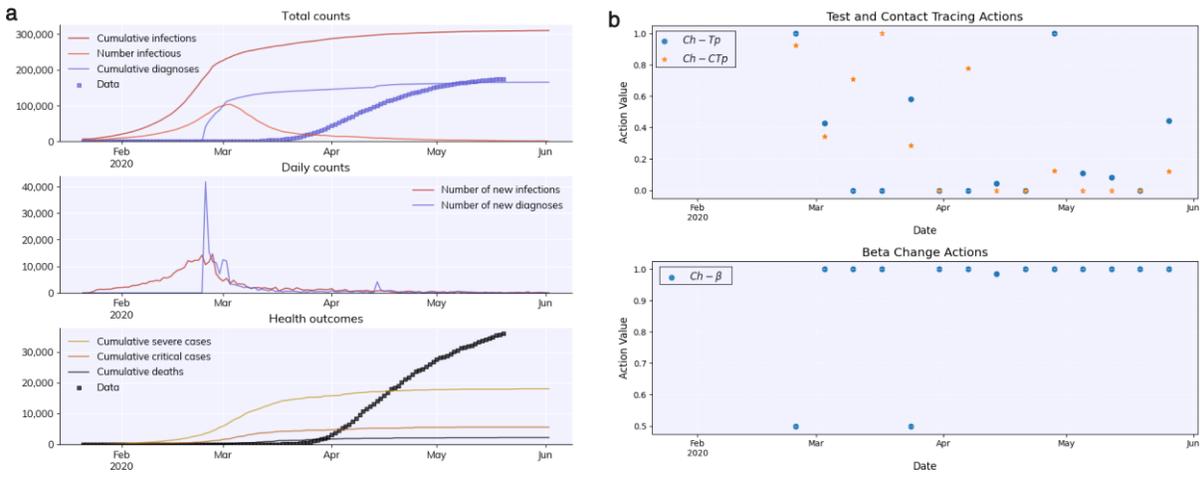

**a** The overall effect diagram of the policy learned by PPO algorithm. **b** The overall effect diagram of the policy learned by PPO algorithm.

In Figure 7, results using the traditional 7-work-7- lockdown strategy in the simulation environment is analyzed. By comparing it with the algorithms used in this section, we find that the strategy learned by agent performs better than 7-work-7-lockdown strategy. The strategy learned by agent effectively controls the epidemic outbreak. On the other hand, the 7-work-7-lockdown strategy doesn't do well in epidemic prevention, as shown in Figure 7. In terms of the economy, as can be seen from Table 1, although 7-work-7-lockdown strategy is superior to the strategy reflecting the real data in terms of controlling the number of infected people, due to the implementation of more stringent lockdown measures, the economic losses have increased. However, the PPO strategy not only reduces the number of infected people but also mitigates economic losses. This highlights the advantages of the strategies studied in this research for dealing with the epidemic and keeping economic balance.

**Fig. 7: Epidemic Development Trends by Using the 7-work-7-lockdown Strategy.**

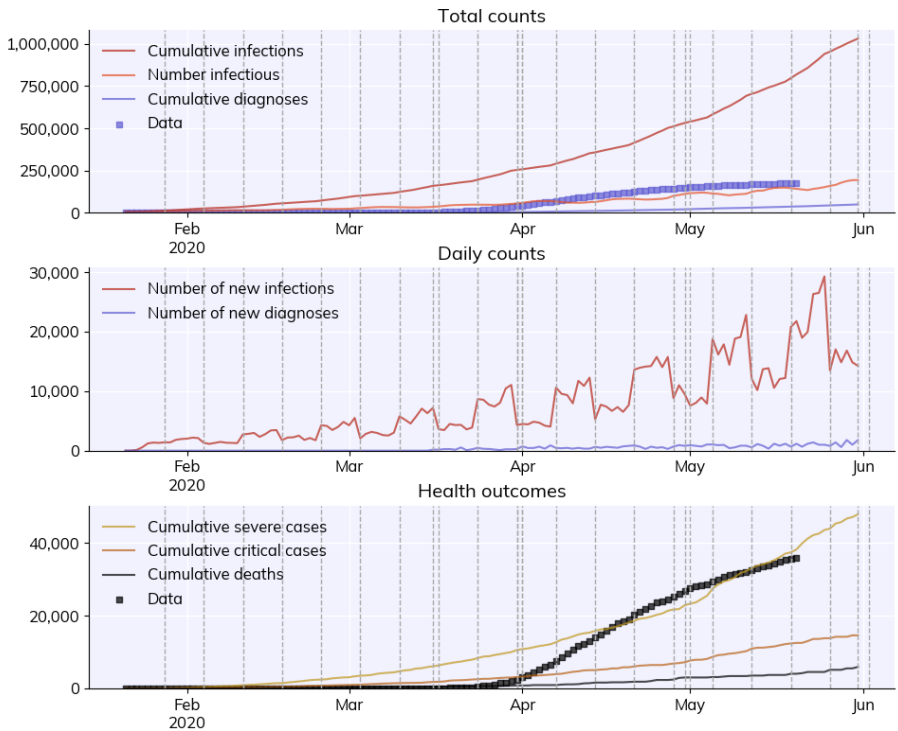

Based on the parameters for fitting real data, 7-work-7-lockdown strategy is imposed, where individuals are allowed to live normally for 7 days and then 80% of them are locked down for 7 days. However, in the lockdown strategy of the agent, at most 50% of individuals will be locked down.

In Figure 8, a comparative study is done on the real-time changes of $R_t$ (real-time reproduction number) under different strategies. According to the basic principles of epidemiology, when the value of $R_t$ is less than 1.0, it means that if the current intervention measures continue, the epidemic is getting better. On the other hand, if $R_t$ is greater than 1.0, it means that the epidemic is getting worse.

It can be noted that in continuous action space, the strategy obtained by PPO algorithm makes the $R_t$ drop quickly below 1.0 because strict containment measures are taken in the early stage. And a series of later intervention measures are carried out to control the value of $R_t$. In the end, the number of infected people is controlled to about 300,000 (see in Figure 6**a**). Under the 7-work-7-lockdown strategy, even though the $R_t$ fluctuates around 1, some epidemic control results are still achieved because control measures are implemented in a timely manner. The final number of infected people reaches about 1,000,000 (as shown in Figure 7). The numbers of total infected individuals under these two strategies are both lower than the strategy fitting real data.

The $R_t$ in the real data fitting strategy drops below 1 at the end of March, while the strategy learned by PPO drops below 1 at the end of February. Just a month's delay causes the cumulative number of infected people to reach 2,900,000 (see in Figure 2). This further shows the importance of taking timely and reasonable intervention measures in infectious disease prevention.

**Fig. 8: Comparison of the Real-time Reproduction Number under Different Strategies.**

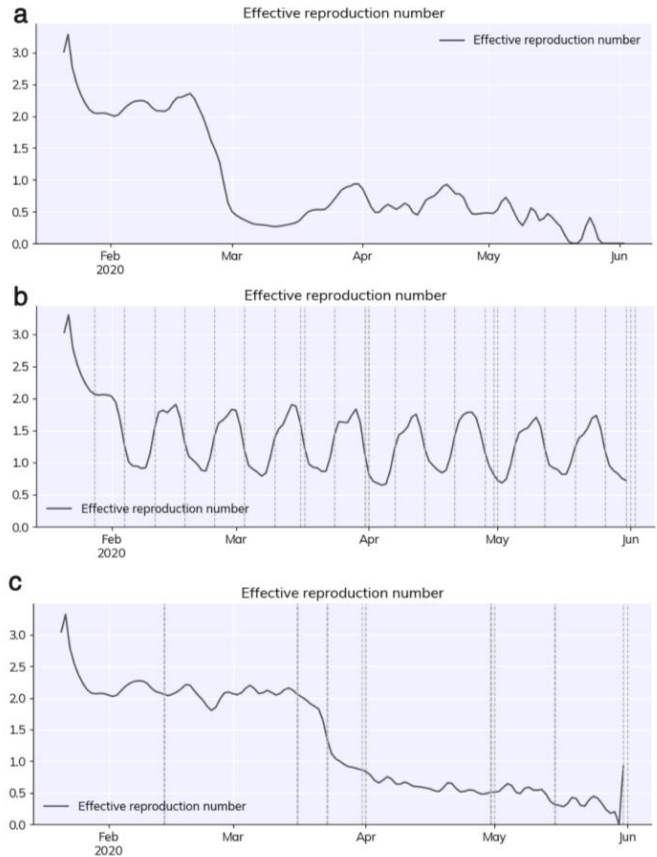

**a** The $R_t$ curve obtained using strategy trained by the PPO algorithm in continuous action space. **b** The $R_t$ curve obtained by using 7-work-7-lockdown strategy. **b** The $R_t$ curve calculated from real-world data.

## Discussion

This study validates the feasibility of using reinforcement learning for infectious disease prevention and control decision-making on an individual agent-based transmission model. Firstly, by using the API provided by the Gym library, we have wrapped the Covasim simulation model into an environment that's good for reinforcement learning. Based on this, value-based DQN algorithm and policy-based PPO algorithm are used to deeply explore the feasibility and effectiveness of their application in different action spaces. Also, a comparative analysis with the traditional 7-work-7-lockdown policy is done. Through experimental studies, we find that using reinforcement learning in this area is practical and has significant advantages compared with traditional lockdown strategy. So, the initial goal of this study is reached, that is, providing an effective decision-making framework for infectious disease prevention and control driven by reinforcement learning.

In addition, the duration of the intervention measures in simulation is studied in this paper, which hasn't been explored before. In reinforcement learning, the idea of time step is very important. Since it's based on the MDP and progresses step by step. If the duration of intervention measures is taken into account, the strategy would become extremely complex. Most previous studies fall into two situations: either they completely ignore the duration and only focus on whether to implement a measure, or they assume that the duration of each intervention measure is fixed (like one day or three days). The drawbacks of these methods are obvious, because the strategies learned often don't match the real situation. Considering that, changes in intervention measures usually don't happen too often, it's more in line with the actual situation to do simulations with one week as a time step. Therefore, this paper sets the simulation time step to seven days.

In the future, we will conduct more in-depth study on the reward function to better balance the effectiveness of epidemic prevention and control with economic considerations.

## **Methods**

### Modification of the Covasim environment

A standardized interface for simulation environments is necessary for reinforcement learning. Considering this, to meet the requirements of reinforcement learning, we must make some changes to the Covasim environment. These changes are set to provide strong support for those working in the field to design and use general algorithms. OpenAI Gym[41], a leading framework in this area, provides standardized Application Programming Interfaces (APIs) and a wide variety of benchmark environments. This feature makes it possible to enclose the Covasim model using Gym's APIs, thus making it easier for reinforcement learning using.

In Covasim, parameters of interventions must be explicitly given before the simulation starts, which is incompatible with the requirements of reinforcement learning. By making specific modifications, the parameters of intervention measures can be adjusted at any point during the simulation.

## Definitions of state space in reinforcement learning

State space is fundamental because it's the key for the agent to understand the environment. The state space must fully and accurately include environmental information. This provides essential data support for the agent to make optimal decisions. The characteristics of the state space can be divided into continuous and discrete types. Each type is especially suitable for specific algorithms.

In this study, how the population is distributed among different infection states is important information that shows the size and situation of the epidemic. The Covasim model uses the SEIRD compartment model. For this research, the number of individuals in different compartments is considered as a discrete state. Since the main goal of this study is to find the best intervention measures, relevant data about interventions, like the number of people tested and those isolated due to testing and contact tracing, are also set as discrete states. In summary, the state space is designed to have eight dimensions: susceptible ($S$), exposed ($E$), infectious ($I$), recovered ($R$), dead ($D$), cumulative tests ($CT$) and cumulative quarantined ($CQ$), and each dimension is a discrete value.

## Definition of action space in reinforcement learning

The action space also needs an assessment of whether it's continuous or discrete. The Covasim model provides a set of well-known intervention measure options. These include lockdown ,in this paper, lockdown is translated into the change of the initial transmission rate ($Ch$-$\beta$ for short, for example, if the value of $Ch$-$\beta$ is 0.6, the new $\beta$ equals 0.6*$\beta initial$, and this means 40 percent of population will be lockdown), the probability of conducting testing procedures (*Ch-Tp* for short), and the probability of using contact-tracing techniques (*Ch-CTp* for short). Both continuous and discrete action space were considered in this paper.

Moreover, we introduce a time delay in the implementation of actions to reflect the challenges of timely epidemic detection because outbreaks are often difficult to be detected before they spread widely in real-life situations. Specifically, the intervention measures will only really activate when the number of infected people reaches a certain scale.

## Definition of reward function in reinforcement learning

In the theoretical framework of reinforcement learning, the reward is crucial feedback through which the agent gets stimuli from the environment. It's very important for evaluating the agent's behavior patterns. In different task situations, the way rewards are distributed can be either sparse or dense. In some cases, non-zero rewards are given only when specific, preset goals are reached. In other cases, the opposite is true, rewards are non-zero for most of the step. To avoid the difficulty of training convergence caused by sparse rewards, the dense reward function has been designed.

Considering that the main goal of this paper is to optimize intervention measures, the reward function mainly includes two parts. The one is health reward ($r_H$), the other is economic reward ($r_E$). These two aspects respectively represent the effectiveness of intervention measures in preventing the spread of epidemics and their impact on economic activities. In addition, when the action space is a continuous one, a penalty ($r_P$) for excessive action changes will be added.

$$r(s_t, a_t, s_{t+1}) = \begin{cases} \lambda_1 r_H + \lambda_2 r_E & \text{if action is discete} \\ \lambda_1 r_H + \lambda_2 r_E + \lambda_3 r_P & \text{if action is continuous} \end{cases}$$

(1)

To make the agent sensitively perceive the practical effects of the intervention measures, $r_H$ is given by:

$$r_H = N_R - \omega_1 N_I - \omega_2 N_S - \omega_3 N_D \qquad (2)$$

Where a positive reward will be given to the new daily number of recovered individuals ($N_R$), while negative rewards will be given to the new daily numbers of infected ($N_R$), severely infected ($N_S$), and dead people ($N_D$). $\omega_1$, $\omega_2$ and $\omega_3$ are coefficients.

The $r_E$ is based on the following assumptions. It is assumed that the economic contribution per day is $C_E$. While infected, quarantined and dead people don't make any economic contribution per day. $r_E$ is given by:

$$r_E = \mu_1 C_E - \mu_2 C_T - \mu_3 C_Q - \mu_4 C_\beta \tag{3}$$

$$C_E = P - M_I - M_Q - M_D \tag{4}$$

$$C_\beta = P(1 - Ch\,\beta) \tag{5}$$

Where P means total population and $M_I$, $M_Q$, $M_D$ means how many people are infected, quarantined, and deceased each day. $C_T$ and $C_Q$ mean economic consumption respectively according to the daily number of people new tested and quarantined. $C_\beta$ means the daily economic consumption if lockdown measure is taken. $\mu_1$, $\mu_2$, $\mu_3$ and $\mu_4$ are economical coefficients.

Due to the difference in scale between $r_E$ and $r_H$, a scaling operation is performed on $r_E$. $r_E$ is also used to calculate economic losses. The original economy refers to the economic level in the situation without the epidemic, calculated by $\mu_1 P$. Daily economic losses are given by:

$$L_E = \frac{\mu_1 P - r_E}{\mu_1 P} \tag{6}$$

$r_P$ is given by:

$$r_P = \begin{cases} -100 \times (|a_t - a_{t-1}| - 0.2) & |a_t - a_{t-1}| > 0.2 \\ 0 & else \end{cases}$$

(7)

This prevents the strategy from fluctuating drastically due to excessive action amplitudes, which would otherwise make it difficult for the training to converge

## Algorithm selection

While selecting reinforcement learning algorithms, comprehensive consideration of multiple factors is necessary. This study focuses on the following key aspects:

Complexity of the environment: The chosen algorithm should match the complexity of the environment, including the size and the dimensional characteristics of both the state space and the action space.

Type of action space: MDP problems need to be solved using different types of reinforcement learning algorithms according to the type of action space.

Sample efficiency and sparse reward Issues: Addressing these challenges requires algorithms that feature intelligent exploration strategies and the ability to effectively utilize samples.

Considering all the factors mentioned above, the following strategy has been adopted for configuring the action space and selecting the reinforcement learning algorithm. When the action space is modeled as a continuous space, the range of *Ch-Tp* and *Ch-CTp* is [0, 1]，the range of *Ch-β* is [0.5, 1], and PPO algorithm is used to make decisions. The lower limit of *Ch-β* is set to 0.5 and it is because a complete lockdown is difficult to sustain in the long term, and a 50% lockdown is easier to implement and maintain. When the action space is modeled as a discrete space, If the boundary is set to 1, the transmission chain of the disease will be cut off, so *Ch-Tp* and *Ch-CTp* can take values in the set {0,0.25,0.50,0.75}, *Ch-β* can take values in the set {0.5, 0.625,0.750,0.875} and DQN algorithm is used to make decision.

## Data Availability
All data of this work can be downloaded from the GitHub repository at [zhangbaida/rlcovasim](zhangbaida/rlcovasim).

## Code Availability
The source code of this work and the trained model can be also downloaded from the GitHub repository at [zhangbaida/rlcovasim](zhangbaida/rlcovasim).

## References

1. WHO. Coronavirus (COVID - 19) Dashboard. https://covid19.who.int (2025).

2. US Centers for Disease Control and Prevention. COVID Data Tracker. https://covid.cdc.gov/covid-data-tracker (2025).

3. Haug, N. et al. Ranking the effectiveness of worldwide COVID - 19 government interventions. *Nat Hum Behav* 4, 1303 - 1312 (2020).

4. Djidjou – Demasse, R., Michalakis, Y., Choisy, M., Sofonea, M. T. & Alizon, S. Optimal COVID - 19 epidemic control until vaccine deployment. MedRxiv, 2020:2020.04.02.20049189 (2020).

5. Yang, B. et al. Effect of specific non-pharmaceutical intervention policies on SARS-CoV-2 transmission in the counties of the United States. *Nat Commun* 12, 3560 (2021).

6. Hellewell, J. et al. Feasibility of controlling COVID-19 outbreaks by isolation of cases and contacts. *Lancet Glob Health* 8, e488-e496 (2020).

7. Kermack, W. O. & McKendrick, A. G. A contribution to the mathematical theory of epidemics. *Proc R Soc Lond A* 115, 700-721 (1927).

8. Grefenstette, J. J. et al. FRED (A Framework for Reconstructing Epidemic Dynamics): an open-source software system for modeling infectious diseases and control strategies using census-based populations. *BMC Public Health* 13, 1-14 (2013).

9. Kerr, C. C. et al. Covasim: an agent-based model of COVID-19 dynamics and interventions. *PLoS Comput Biol* 17, e1009149 (2021).

10. AlMahamid, F. & Grolinger, K. Reinforcement learning algorithms: An overview and classification. *2021 IEEE Canadian Conference on Electrical and Computer Engineering (CCECE)*, 1-7 (2021).

11. Kuzmin, V. Connectionist Q-learning in robot control task. *Proc Riga Tech Univ* 112, 112-121 (2002).

12. Fujimoto, S., Hoof, H. & Meger, D. Addressing function approximation error in actor-critic methods. *Proc Int Conf Mach Learn, PMLR* 1587-1596 (2018).

13. Anwar, A. & Raychowdhury, A. Autonomous navigation via deep reinforcement learning for resource constraint edge nodes using transfer learning. *IEEE Access* 8, 26549-26560 (2020).

14. Mnih, V. et al. Human-level control through deep reinforcement learning. *Nature* 518, 529-533 (2015).

15. Lillicrap, T. P. et al. Continuous control with deep reinforcement learning. Preprint at https://arxiv.org/abs/1509.02971 (2015).

16. Schulman, J., Wolski, F., Dhariwal, P., Radford, A. & Klimov, O. Proximal policy optimization algorithms. Preprint at https://arxiv.org/abs/1707.06347 (2017).

17. Pathak, D., Agrawal, P., Efros, A. A. & Darrell, T. Curiosity-driven exploration by self-supervised prediction. *Proc Int Conf Mach Learn, PMLR* 2778-2787 (2017).

18. Kulkarni, T. D., Narasimhan, K., Saeedi, A. & Tenenbaum, J. Hierarchical deep reinforcement learning: Integrating temporal abstraction and intrinsic motivation. *Adv Neural Inf Process Syst* 29 (2016).

19. Haarnoja, T., Zhou, A., Abbeel, P. & Levine, S. Soft actor-critic: Off-policy maximum entropy deep reinforcement learning with a stochastic actor. *Proc Int Conf Mach Learn, PMLR* 1861-1870 (2018).

20. Karin, O. et al. Adaptive cyclic exit strategies from lockdown to suppress COVID-19 and allow economic activity. medRxiv 2020, 2020.04 (2020).

21. Huberts, N. F. D. & Thijssen, J. J. J. Optimal timing of non-pharmaceutical interventions during an epidemic. *Eur J Oper Res* 305, 1366-1389 (2023).

22. Aleta, A. et al. Modelling the impact of testing, contact tracing and household quarantine on second waves of COVID-19. *Nat Hum Behav* 4, 964-971 (2020).

23. Ash, A., Bento, A. M., Kaffine, D., Rao, A. & Bento, A. I. Disease-economy trade-offs under alternative epidemic control strategies. *Nat Commun* 13, 3319 (2022).

24. Kaleta, M. et al. Long-term spatial and population-structured planning of non-pharmaceutical interventions to epidemic outbreaks. *Comput Oper Res* 146, 105919 (2022).

25. Padmanabhan, R., Meskin, N., Khattab, T., Shraim, M. & Al-Hitmi, M. Reinforcement learning-based decision support system for COVID-19. *Biomed Signal Process Control* 68, 102676 (2021).

26. Kwak, G. H., Ling, L. & Hui, P. Deep reinforcement learning approaches for global public health strategies for COVID-19 pandemic. *PLoS One* 16, e0251550 (2021).

27. Song, S., Zong, Z., Li,Y., Liu, X. & Yu, Y. Reinforced epidemic control: Saving both lives and economy. Preprint at https://arxiv.org/abs/2008.01257 (2020).

28. Libin, P. J. et al. Deep reinforcement learning for large-scale epidemic control. *Lect Notes Comput Sci, Springer* 155-170 (2021).

29. Ma,i A., Gupta, N., Abouzied, A. & Shasha, D. Planning multiple epidemic interventions with reinforcement learning. Preprint at https://arxiv.org/abs/2301.12802 (2023).

30. Bampa, M., Fasth, T., Magnusson, S. & Papapetrou, P. EpidRLearn: Learning Intervention Strategies for Epidemics with Reinforcement Learning. *Int Conf Artif Intell Med, Springer* 189-199 (2022).

31. Feng, T., Song, S., Xia, T. & Li, Y. Contact tracing and epidemic intervention via deep reinforcement learning. *ACM Trans Knowl Discov Data* 17, 1-24 (2023).
21

32. Feng, T. et al. Precise mobility intervention for epidemic control using unobservable information via deep reinforcement learning. *Proc ACM SIGKDD Int Conf Knowl Discov Data Min* 2882-2892 (2022).

33. Ohi, A. Q., Mridha, M. F., Monowar, M. M. & Abdul Hamid, M. d. Exploring optimal control of epidemic spread using reinforcement learning. *Sci Rep* 10, 22106 (2020).

34. Hasselt, H. Double Q-learning. *Adv Neural Inf Process Syst* 23 (2010).

35. Kompella, V. et al. Reinforcement learning for optimization of COVID-19 mitigation policies. Preprint at https://arxiv.org/abs/2010.10560 (2020).

36. Kerr, C. C. et al. Controlling COVID-19 via test-trace-quarantine. *Nat Commun* 12, 2993 (2021).

37. Panovska-Griffiths, J. et al. Determining the optimal strategy for reopening schools, the impact of test and trace interventions, and the risk of occurrence of a second COVID-19 epidemic wave in the UK: a modelling study. *Lancet Child Adolesc Health* 4, 817-827 (2020).

38. Pham, Q. D. et al. Estimating and mitigating the risk of COVID-19 epidemic rebound associated with reopening of international borders in Vietnam: a modelling study. *Lancet Glob Health* 9, e916-e924 (2021).

39. Bushaj, S., Yin, X., Beqiri, A., Andrew, D. & Büyüktahtakın, IE. A simulation-deep reinforcement learning (SiRL) approach for epidemic control optimization. *Ann Oper Res* 328, 245-277 (2023).

40. Akiba, T., Sano, S., Yanase, T., Ohta, T. & Koyama, M. Optuna: A next-generation hyperparameter optimization framework. *Proc ACM SIGKDD Int Conf Knowl Discov Data Min* 2623-2631 (2021).

41. Brockman, G. et al. OpenAI Gym. Preprint at https://arxiv.org/abs/1606.01540 (2016).

## Acknowledgements

This project has received support from the National Natural Science Foundation of China and thanks for the suggestions provided by professor Chengli Zhao.

## Ethics declarations

Competing interests

The authors declare no competing interests.

## Supplementary Information

Tables

**Table 1 Economic Losses of Different Strategies. PPO strategy is learned in continuous action space**

| Strategy | Economic Loss |
| --- | --- |
| PPO strategy | 10.25% |
| 7-work-7-lockdown strategy | 38.01% |
| Reflecting real-world strategy | 36.61% |

**Table 2 Parameters of PPO. The parameters in the discrete action space and the continuous action space are the same.**

| Parameter | Value |
| --- | --- |
| n_steps | 190 |
| batch_size | 19 |

| learning_rate | 0.0001 |
| n_epochs | 10 |
| gamma | 0.99 |
| clip_range | 0.2 |

"n_steps" determines the number of steps in which the agent interacts with the environment each time data is collected.

After collecting data for "n_steps", these data will be divided into multiple mini-batches, and the number of samples in each mini-batch is "batch_size".

"learning_rate" controls the step size of model parameter updates.

"gamma" is the discount factor, which is used to calculate the present value of future rewards.

"clip_range" is used to limit the amplitude of policy updates in the PPO algorithm..

**Table 3 Parameters of DQN algorithm with PER**

| Parameters of the DQN algorithm ||
|---|---|
| **Parameter** | **Value** |
| buffer_size | 1900 |
| batch_size | 19 |
| learning_starts | 57 |

| learning_rates | 0.0001 |
|---|---|
| target_update_interval | 95 |
| tau | 1 |
| gamma | 0.99 |
| **Parameters of PER** | |
| **Parameter** | **Value** |
| alpha | 0.6 |
| beta | 0.4 |
| beta_increment_per_sampling | 0.001 |

"buffer_size" is the size of the experience replay buffer.

"learning_starts" means the number of pieces of experience data that need to be collected before the formal network training begins.

"target_update_interval" is the update interval of the target network.

"tau" is the soft update coefficient, which is used for the soft update mode of the target network.

"alpha" controls the priority weight in the prioritized experience replay.

"beta" is used to correct the bias caused by non-uniform sampling in the prioritized experience replay

"beta_increment_per_sampling" is the increment of beta for each sampling.

**Table 4 Covasim Parameter Settings for Fitting Real Data**

| Parameter | Value |
| --- | --- |
| total_pop | 67.86e6 |
| pop_size | 10000 |
| pop_scale | total_pop / pop_size |
| pop_type | hybrid |
| pop_infected | 5856 |
| n_days | 133 |
| beta($\beta_{initial}$) | 0.005997 |
| contacts | h:3.0, s:20, w:20, c:20 |
| asymp_factor | 2 |
| sus_ORs[0,10] | 1 |
| sus_ORs[0,20] | 1 |

The population data are presented in the unit of "person"

"beta($\beta_{initial}$)" represents the initial transmission rate between individuals.

"contacts" represents the number of individuals contacted by individuals of different space types.

sus_Ors[X,Y] represents the odds ratios for relative susceptibility of the age group from X to Y.